\documentclass[letterpaper, 10 pt, conference]{ieeeconf}  

\IEEEoverridecommandlockouts                              
\usepackage{graphicx} 
\graphicspath{ {./images/} }
\title{\LARGE \bf
Metrics for the Evaluation of localisation Robustness
}

\author{Siqi Yi, Stewart Worrall and Eduardo Nebot*
\thanks{*Authors are from Australian Center for Field Robotics, the University of Sydney, Australia}
}

\begin{document}

\maketitle
\thispagestyle{empty}
\pagestyle{empty}

\begin{abstract}
 Robustness and safety are crucial properties for the real-world application of autonomous vehicles. One of the most critical components of any autonomous system is localisation. During the last 20 years there has been significant progress in this area with the introduction of very efficient algorithms for mapping, localisation and SLAM. Many of these algorithms present impressive demonstrations for a particular domain, but fail to operate reliably with changes to the operating environment. The aspect of robustness has not received enough attention and localisation systems for self-driving vehicle applications are seldom evaluated for their robustness. In this paper we propose novel metrics to effectively quantify localisation robustness with or without an accurate ground truth. The experimental results present a comprehensive analysis of the application of these metrics against a number of well known localisation strategies. 
\end{abstract}


\section{INTRODUCTION}

Robustness can be defined as the ability of a system to tolerate perturbations, or the capability to endure uncertainty and noise, before system assumptions are irreversibly violated. Vehicle localisation is a critical function of an autonomous driving system and consequently requires a very high standard of integrity, robustness and stability. Significant research efforts have been dedicated to improving the accuracy of localisation algorithms for a specific domain, but in the authors' view, there is not enough emphasis on robustness. For example, there has been a significant focus on improving the localisation performance using the KITTI odometry dataset\cite{Geiger2012CVPR}, where the best-performing algorithm in translational accuracy achieved an impressive 0.57\% \cite{zhang2015visual}. This is an improvement of 0.73\% difference from the algorithm at 50th place, with only 0.0146\% improvement per algorithm in the top 50 algorithms. These efforts has been concentrated on improving accuracy in a single domain without a consideration for generalisation to other domains.

Localisation based on visual odometry/SLAM  has made significant progress during recent years.  These algorithms can be classified as high accuracy - low robustness systems. They rely on the validity of assumptions such as small changes to lighting, no shadows, low angular and linear vehicle velocity, an abundance of features, and no moving objects in camera field of view. Operating in an urban road environment results in frequent violations of these assumptions which can lead to localisation failure. Moreover, because visual features are unstable and transient, it is hard to maintain a map of features that can be used for localisation purposes. Failure modes need to be carefully characterized to improve robustness by incorporating mechanisms to recover from becoming lost.

There is an inherent difficulty of the quantitative evaluation of localisation and maps: ground truth is hard to obtain for vehicle trajectories and maps generated from SLAM algorithms. RTK-GPS and motion capture cameras are conventional sensors that have enough accuracy of 2cm or less and are generally trusted as sources of ground truth. These measurements however are not generally applicable to an urban road environment due to a high proportion of areas suffering from satellite multipath due to buildings and trees, and satellite-denied areas such as tunnels and indoor car parks. Accuracy metrics such as \cite{Steder2009} and \cite{Sturm2012} are not suited to this situation as they have a hard constraint on needing a high accuracy localisation source as a benchmark for calculations. When designing robustness metrics for realistic urban datasets, we must take into account that most of the time a high accuracy ground truth is not available. Our proposed metrics does not require ground truth, as we evaluate the map robustness in conjunction with localisation.

The paper is structured as follow: First, we present the related work in section \ref{relatedwork} and introduce localisation robustness metrics in section \ref{metrics}. Section \ref{system} presents a description of how the globally consistent maps are created from our own datasets, and the implementation of localisation systems based on GPS/GNSS, lidar and IMU. The results of applying the novel robustness metrics are presented in section \ref{result}.


\begin{figure}
    \centering
    \includegraphics[width=\columnwidth]{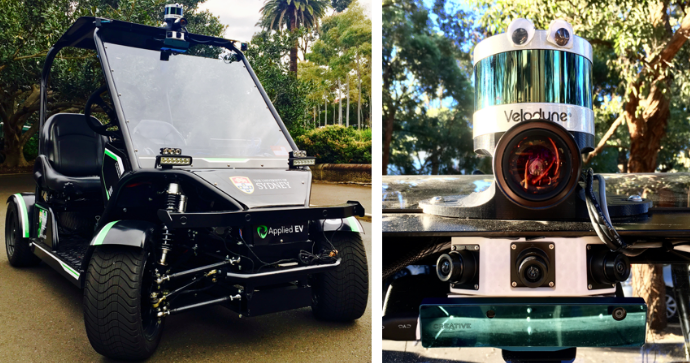}
    \caption{One of the electric vehicles used for online localisation testing and data acquisition}
    \label{fig:sample_figure}
\end{figure}

\section{RELATED WORK} \label{relatedwork}
There is existing work looking at localisation quality from a variety of domains including vision, lidar and GPS based localization. One common feature of the previous work is that they focus on a specific set of sensors, and are not designed to compare different localisation approaches.

\cite{Merzic2017} proposed an algorithm for computing visual map quality. 
This method is able to predict localisation quality at a given map query point. It is not a metric capable of gauging robustness in experimental data and does not generalizes to many localisation systems and sensors. Map quality in \cite{Merzic2017} is an indicator or prediction for localisation performance. The value varies with the parameterisation or weighting of the evaluation function, and its usage is restricted to visual landmark based localisation.

\cite{Linegar2016} claim to have designed a visual localisation system that is can be operated at any time of the day or night, and in all weather conditions. The robustness metric they used is the portion of localisation failure in each specific dataset. Localisation failure is defined as not receiving an observation for more than 20m.
This concept of robustness can be too restrictive as different localisation algorithms can have different failure modes. 

\cite{Porav2018} showed the robustness performance of their algorithms using a similar measure of probability of absence of updates metric proposed by our paper. This paper does not however provide a detailed explanation or focus on this particular metric.

Self-assessing localisation \cite{Manuel2018} is a stable lidar feature ICP map matching localisation system that has the capability to self-assess consistency online. Normalized Innovation Squared (NIS) is used for the metric of a consistency test for spatial uncertainty.
Clutter rate and detection probability are also tested for consistency, but they are metrics specific to discrete features modeled by Random Finite Sets multi-object tracking. 

There is a considerable body of research focusing on improving visual localisation robustness due to the brittle nature of visual features with time of the day, lighting, or other environmental variables. One example of this research is \cite{Burki2016}. In this work, they design a ranking function and assign a quality rank for each map feature so that an adaptive feature selection policy can be enacted according to variation in the appearance of features due to events such as time of day or night. \cite{zhang2015visual} tested robustness against fast motion and undesirable lightening conditions. They concluded robustness and accuracy is a trade-off when using fish-eye cameras, and improved robustness by incorporating multi-sensor fusion. \cite{Mostegel2014} proposed geometric point quality, point recognition probability and localisation quality as robustness metrics, and presented methodology specific to visual features and localisation. 

In the field of GNSS localisation, \cite{Worrall2015} suggested validation of GPS signal with IMU as a measure for GPS integrity, since IMU errors are relatively constant and GPS errors are strongly influenced by environment.  

Utilising datasets with accurate and absolute ground truth, \cite{Steder2009} and \cite{Sturm2012} proposed relative displacement/pose error (RPE), including relative translation and rotation errors, and absolute trajectory error (ATE) as accuracy metrics that provide a framework for evaluating and comparing SLAM in large public datasets. Due to lack of ground truth, \cite{Smedberg2017} tested localisation repeatability as a localisation performance metric in which the variance of position estimates from multiple robots moving along the same trajectories were compared. Similarly, \cite{Strasdat2010} evaluated algorithms by implementing feature drop out, as ground truth cannot be measured. \cite{Schroeter2008} evaluated robustness of bearing only SLAM by considering rotational and translational error with respect to simulations of variable landmark density, level of the rate of incorrect data association and compass noise.

\section{LOCALISATION ROBUSTNESS METRICS} \label{metrics}
In this section, we introduce two metrics that can effectively characterize the robustness of localisation. In section \ref{result} we demonstrate their practical usage when evaluating the performance of localisation algorithm with real-life datasets.

\subsection{Valid prior threshold (VPT)} \label{Valid prior threshold}
\begin{figure}
    \centering
    \includegraphics[width=\columnwidth]{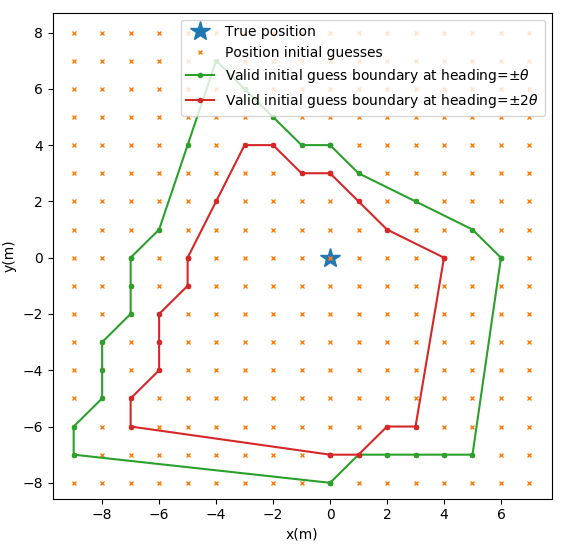}
    \caption{An example of computing boundary of valid prior threshold at one map location}
    \label{fig:illust_icpboundary}
\end{figure}
Localisation against a map requires an initial pose estimate to reduce the search space and resolve ambiguity for data association.
This can be challenging on initialisation of the algorithm, and after a period without observations when the uncertainty of the pose prior is the highest.
The robustness of map based localisation can be measured by how tolerant the matching process is to uncertainty in the pose prior. This is dependent on many factors such as the density, or ambiguity in the distribution of the landmarks. 
We define here a valid prior threshold (VPT) metric as a search over a map location to test the tolerance to more uncertain priors, which can be used to indicate the likelihood that a map matching observation will be correct for a given location. 
The boundary of VPT is determined by performing a search over the state variables, which in our case include x, y and heading as shown in Fig \ref{fig:illust_icpboundary}.
The size of the boundary will be smaller if there is a poor distribution of features, indicating that is can also be used as a metric for map quality.

Matching algorithms are used to determine pose by matching the output of sensors such as cameras or lidars to an existing map that contains a spatial representation of pointclouds, images, or some other high level features. The output pose and transformation estimate from the matching process can then be incorporated into a sensor fusion algorithms to produce a state estimate. The initial estimate of the pose is required as the input to the matching algorithms.
During the filtering process this inital pose estimate comes from the filter state, or initialisation of the filter. 
When state estimates are far away from vehicle true states and are fed to matching algorithms as initial guesses, matching algorithms fail to determine poses, or output wrong poses.


Fig. \ref{fig:illust_icpboundary} illustrates one example implementation of valid prior threshold. At time t the lidar sensed a constellation of features at the true vehicle pose A indicated by the blue star. We position a fixed coordinate system origin at pose A, and perform feature matching as if the lidar processing algorithm perceived the constellation of features at each of the orange positions, and as if vehicle heading was \(\pm\theta\). The figure indicates that at all the orange dots with \(-\theta \leq heading \leq \theta\) within the green boundary the algorithm is able to evaluate the true vehicle pose. Within this area, the algorithm is able to incorporate the observation and update the vehicle pose. Beyond the boundary algorithm will not be able to do the proper matching to incorporate the observations. The same process are repeated to find boundaries at \(\pm2\theta\), \(\pm3\theta\)... to form a contour graph, see section \ref{prior boundary result} and Figure \ref{fig:icpBoundPerFrame} for dataset evaluation details.

\begin{figure}
    \centering
    \includegraphics[width=\columnwidth]{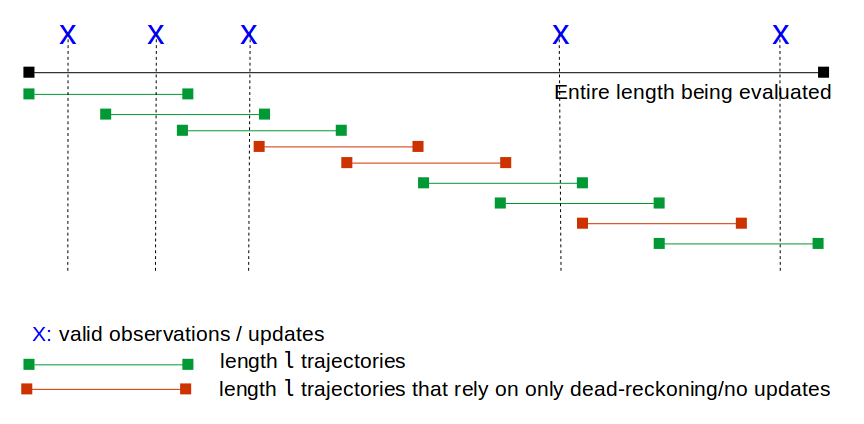}
    \caption{An example of computing probability of absence of updates(PAU) of length \(l\). PAU = 0.33 in this case. }
    \label{fig:illust_probDR}
\end{figure}

\subsection{Probability of Absence of Updates(PAU)} \label{Probability of dead-reckoning}
There are two main coordinate frames in which localisation takes place, global and relative. Observations in a global reference frame rely on a sensor that can measure some known feature from the environment, such as lidar or image based features from a prior map or GNSS satellite positions. Relative motion can be observed using a range of sensors such as an inertial measurement unit (IMU), wheel encoders or through visual/lidar based odometry. Observations in a global reference frame are determined by the quality and density of features in the environment. Relative motion does not rely on a set of known global features, and through IMU/encoder based observations can update at a high frequency. 


A common localisation strategy is to use a statistical filter to estimate the absolute position and heading (and other state variables) of a system. To estimate the pose in a global reference frame, the filtering process will generally involve relative motion in the prediction step to estimate the change in global pose between update steps obtained through observations in the global reference frame. A typical example of this process is using a combination of high frequency IMU/encoder information to predict the change in global pose observed by a GNSS sensor. The relative motion is generally smooth and locally consistent allowing global updates to be potentially rejected if they are statistical outliers. The relative motion is referred to as dead-reckoning, and in the absence of other information can be estimated using a constant velocity or constant acceleration model.


Dead-reckoning accumulates error and drifts relative to the time and distance traveled. After long distances, an absence of global pose updates leads to potentially significant dead-reckoning error with an associated increase in the uncertainty of the pose.
During normal operation, the uncertainty is bounded by regular global position updates to improve the pose estimate. 
As the pose prior is used for map based localisation, if the uncertainty grows too large it can end up with a pose that is outside the range of the valid prior threshold described in the previous section. 
This can potentially lead to catastrophic failure of the filter, and so the regularity of the global observations is considered a measure of robustness for the localisation algorithm.

To generate a metric that can encapsulate the regularity of the global observations used for updating the filter, we define a measure of probability of absence of updates (PAU) as follows.
Let \(TRAJ_l\) denote the set of all trajectories of a given length \(l\) a vehicle is able to travel in a given area, as shown by green and red line segments in Figure \ref{fig:illust_probDR}. Let \(AU_l \subseteq TRAJ_l\) represents the subset of trajectories that are not able to get updates from other sensors and map sources for the length \(l\), or only rely on dead-reckoning, as shown by red line segments. The PAUs of length \(l\) can be computed by:
\[P(AU_l) = \frac{\mid AU_l \mid}{\mid TRAJ_l \mid} \]
where \(\mid . \mid\) is the number of elements in a set. In the illustration shown in Fig. \ref{fig:illust_probDR}, \(P(AU_l) = \frac{3}{9} = 0.33\). Plotting \(P(AU_l)\) against \(l\) produces a PAU curve indicates the likelihood of receiving global updates within a given area, and gives us a sense how much reliance is put on dead-reckoning to restrict uncertainty growth and prevent localisation failure. Fig \ref{fig:probsLostQuad} shows examples of this curve.  The PAU curve decreases with \(l\) as it is inversely related to the length of the measured trajectory. For a short trajectory length there is a higher probability not receiving any valid updates because the global sensor updates are generally lower frequency and, depending on the environment, affected by noise. The tail of the curves represents longer trajectories without receiving global pose updates, indicating an accumulation of dead-reckoning error potentially resulting in the loss of localisation. 


This metric is very useful for comparative analysis of the distribution of different types of global pose observations. By drawing PAU curves on the same plot, localisation robustness can be quantified as the area below a curve, improvements can be measured by the area between the curves. The combination of global pose updates produces lower curves exhibiting improved robustness performance. This is because for the same trajectory length a lower curve has a reduced chance of not receiving any updates.
This is an excellent tool to determine the reliability of receiving pose updates given a set of sensors and features.
See section \ref{Probability of dead-reckoning result} for more information.

\section{SYSTEM DESCRIPTION} \label{system}
We developed and tested the metrics described in this paper using data collected from our autonomous vehicle platforms. The metrics serve as part of a localisation framework that can be easily extended to evaluate the robustness of other feature based localisation systems.
\subsection{Hardware Setup}
Our two electric vehicles at the University of Sydney are equipped with a variety of sensors. The localisation system runs constantly on both vehicle and is able to operate in real time. The current localisation system is based on lidar combined with other sensors such as IMU, wheel encoders and GPS.

\subsection{Evaluation Dataset}
Datasets were collected at the main campus of the University of Sydney along a prescribed route every week during different times of day, weather and seasons. Multiple drives in several particular areas of interest for localisation and mapping were also recorded for the development and testing of algorithms.

\subsection{Lidar features selection}
An algorithm was developed to extract pole and building corner features from a lidar point cloud. These are two types of feature that can be reliably detected by lidars and are abundant in an urban environment. Poles are defined as cylindrical objects that are separated from other structures. We constrain the definition of a poles to be tall and thin in order to reduce false detections from non-pole objects such as pedestrians, which can be similar shape within a lidar point cloud. Building corners are defined as the intersection of two planes/walls with a measured angle. Both poles and corners make robust lidar features as they are generally orthogonal to plane in which the vehicle operates such that their projection into 2 dimensions can be represented as point features on the map. Poles and corners are high level semantic features that are intuitive for both human and computer algorithms. They are most often persistent and distinguishable over long time frames. The seasons, illumination, time of year and day do not drastically affect the reliability of detection. Our experiments over time have demonstrated that these features are suitable for precise localisation in our test environment. 
We use these features only as an example to determine the quality metrics. This process can be applied to any set of global pose updates.

\subsection{Stable feature ICP matching}
We use an Iterative Closest Point (ICP) algorithm to associate an observed set of lidar features against an existing map for localisation, with a prior feature map created in a separate process. An initial estimate of vehicle pose within the map coordinate system is needed for starting the localisation process, and this pose prior is provided by GPS during initialisation. During the filtering process, subsequent pose priors for initialising the ICP algorithm are taken from the filter state which relies on the prediction updates from IMU/encoder based dead reckoning.

\begin{figure}
    \centering
    \includegraphics[width=\columnwidth]{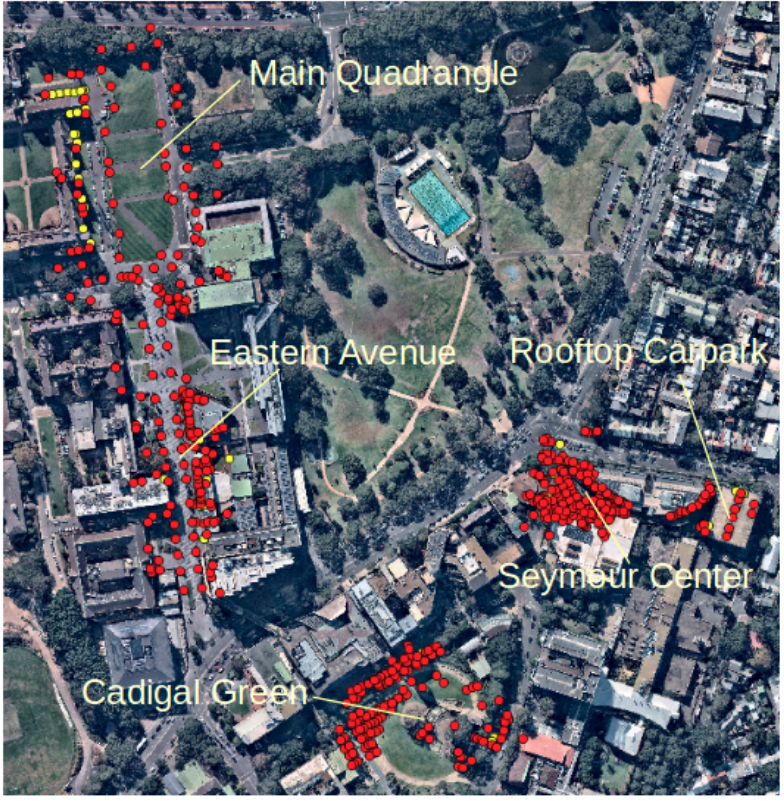}
    \caption{Map features on Google earth satellite image. Red: poles; Yellow: corners}
    \label{fig:map}
\end{figure}
\begin{figure}
    \centering
    \includegraphics[width=\columnwidth]{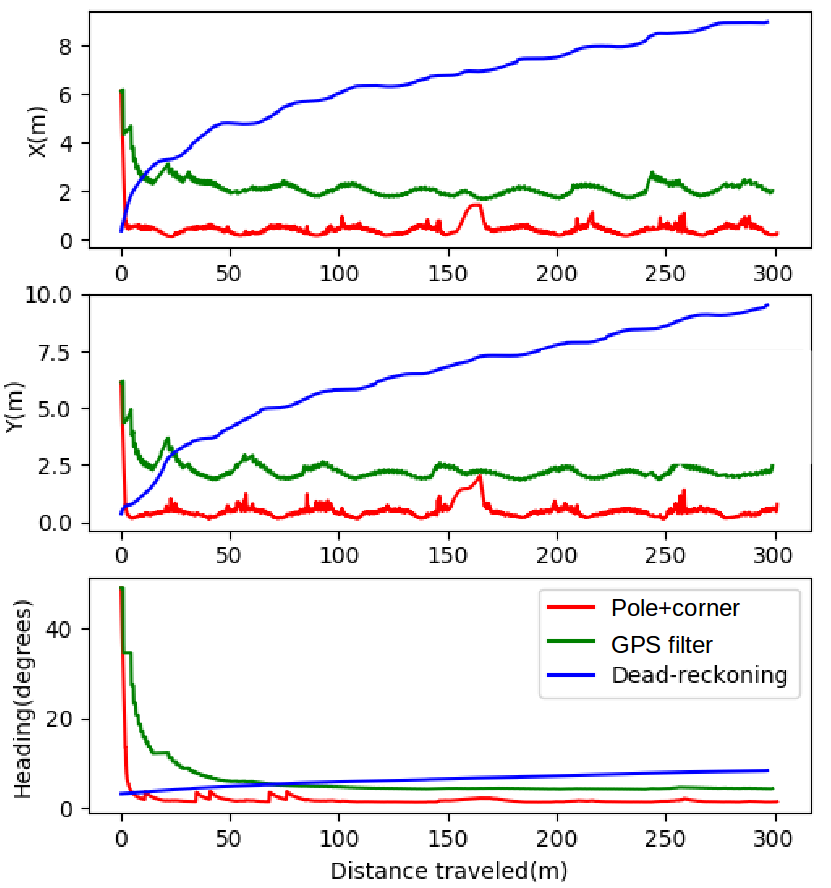}
    \caption{Seymour dataset 95\% confidence bound. X, Y, Heading are in UTM frame. Pole+corner, GPS and dead reckoning localisation modes are represented in red, green and blue respectively.}
    \label{fig:seymourUncerty}
\end{figure}
\begin{figure}
    \centering
    \includegraphics[width=\columnwidth]{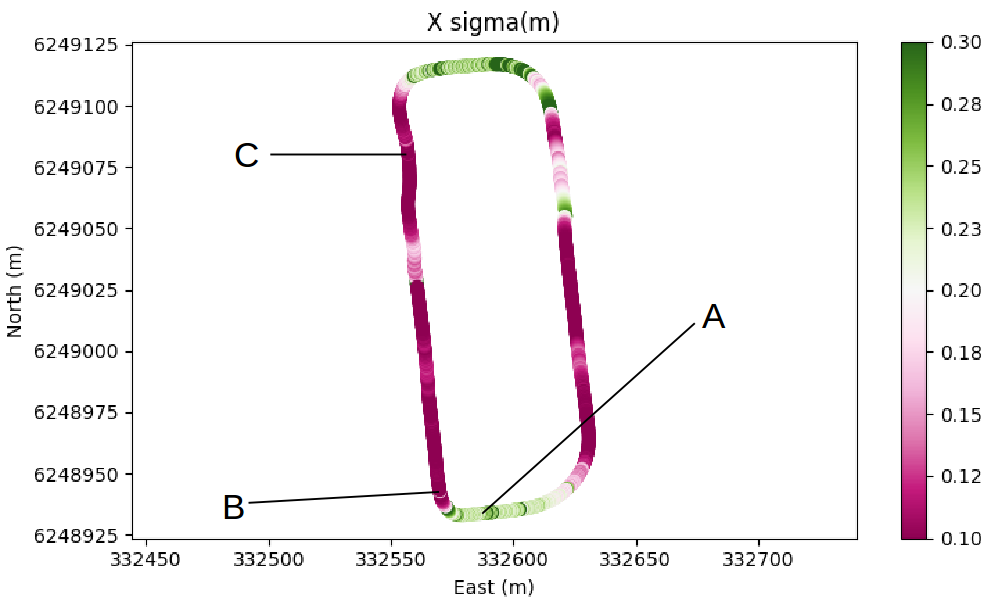}
    \caption{Trajectory mean in UTM frame of Main Quadrangle dataset from pole+corner localisation. Color denotes standard deviation in Eastings. A, B and C are locations where valid prior boundary are evaluated in Fig. \ref{fig:icpBoundPerFrame}.}
    \label{fig:pathwithxsigma}
\end{figure}

\begin{figure}
    \centering
    \includegraphics[width=\columnwidth]{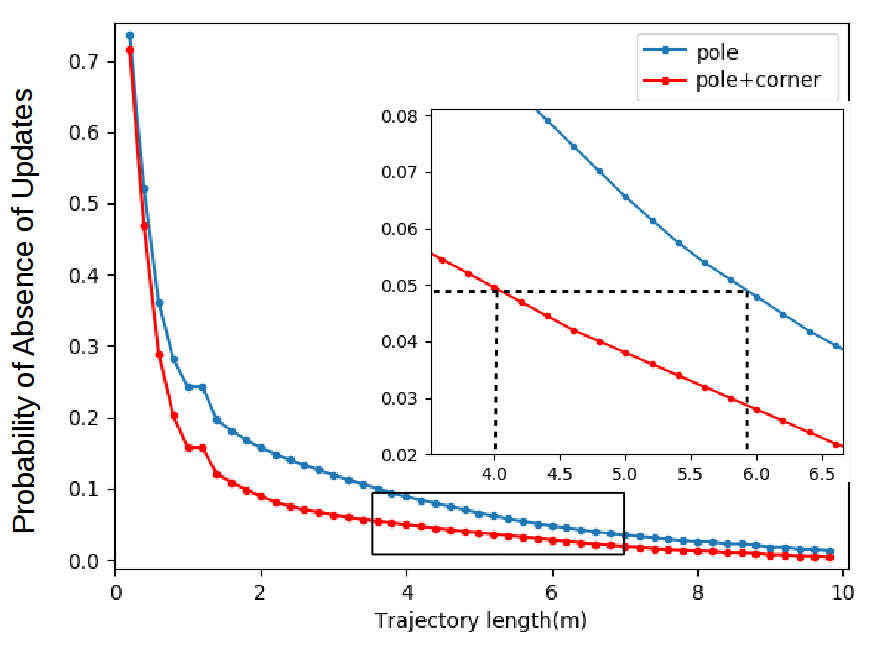}
    \caption{probability of absence of updates curves at Main Quadrangle. Blue curve is evaluated from pole+corner localisation, red curve is evaluated from pole localisation mode. An enlarged graph shows at PAU=0.05, pole absence of updates length is 4m, pole+corner is around 5.9m}
    \label{fig:probsLostQuad}
\end{figure}
\begin{figure}
    \centering
    \includegraphics[width=\columnwidth]{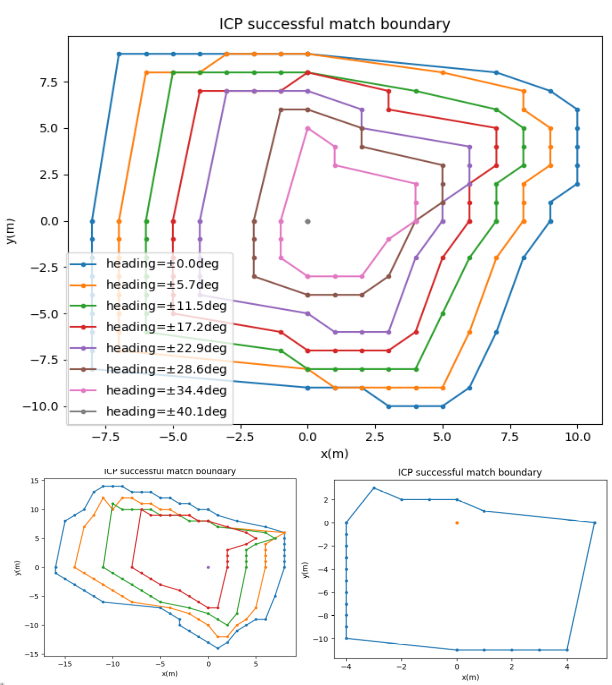}
    \caption{Valid prior threshold boundaries at 3 locations in Main Quadrangle. Top, bottom left, bottom right is evaluated at location A, B and C shown in Figure \ref{fig:pathwithxsigma}}
    \label{fig:icpBoundPerFrame}
\end{figure}
\begin{figure*}
    \centering
    \includegraphics[width=\textwidth, height=4cm]{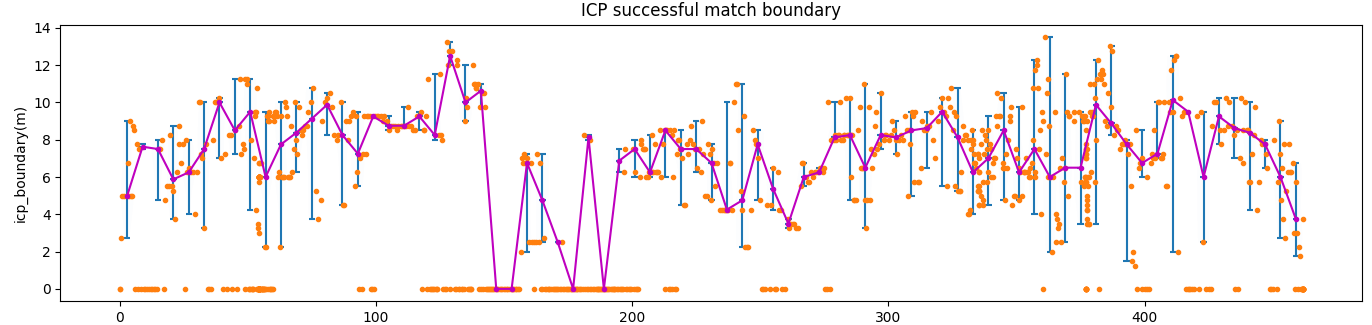}
    \caption{VPT boundary radius evaluated at \(\theta\) = 5.7 degrees (0.1 radian) along the whole length of Main Quadrangle loop. Orange points are prior boundary radius computed from ICP matching all lidar frame in Main Quadrangle dataset with map. If no match can be find or no features are detected, the radius is plotted at zero. Blue lines are the span of non-zero radius of for every 6m along the loop. Magenta line connects the median radius of the 6m stretches.}
    \label{fig:icpBoundAll}
\end{figure*}
\begin{figure*}
    \centering
    \includegraphics[width=\textwidth, height=4cm]{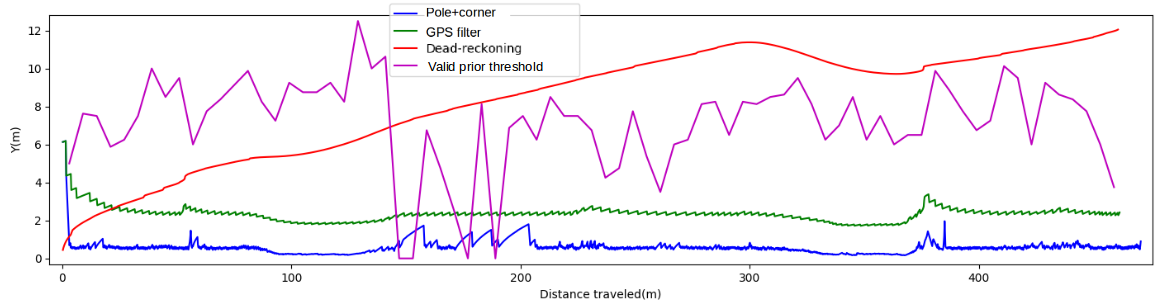}
    \caption{VPT boundary radius(magenta) plotted with 95\% confidence bound of pole+corner(blue), GPS(green) and dead-reckoning(red) filters. Careful analysis of this figure reveals robustness of the algorithms along the 470m length.}
    \label{fig:allsensorEval}
\end{figure*}
\subsection{Map Creation}
A lidar feature map was created from datasets collected from several different areas of the University of Sydney Camperdown campus including a) the main quadrangle that is a 470m loop consisting of several sections of tall buildings, one section with very few features, crowds of pedestrians, slopes and uneven road surfaces b) Eastern Avenue which is a 250m straight pedestrian road surrounded by buildings c) Cardigal Green with rows of evenly distributed trees d) Seymour Center that is closely surrounded by three buildings, some of which are glass buildings. Glass buildings were particularly interesting for extracting lidar features because the window/glass dividers are detected as pole features because glass does not usually reflect laser light and is perceived as empty space e) Rooftop of engineering carpark where RTK-GPS signals are available throughout with 2cm accuracy. The rooftop carpark is seven stories tall, and was the only area capable of receiving RTK-GPS as a ground truth. With the variety of environments captured in the datasets, we are able to obtain a deep insight into the differences in localisation performance for each type of area.

Whilst our mapping approach is capable of online map creation, we create maps offline as a separate process from localisation. For the purpose of localisation in an autonomous vehicle application, we require a global pose estimate against a known map. The popular SLAM paradigm is capable of simultaneously creating a map and positioning the robot, and is useful for operation in an unknown environment. This is also useful when map features changes too rapidly with respect to time, illumination or viewing angle. Our use case of autonomous driving does not include the exploration of unknown environments while driving as it imposes a risk to passenger and vehicle safety, and is contrary to the goal of driving to a specific location.

We have experimented with using both EKF-SLAM and graph SLAM to create the feature maps using identical features and datasets. Graph slam was performed using local bundle adjustment over most recent 3 frames of lidar features. IMU/Encoders are used to compute the transformations between the 3 vehicle poses. Graph optimization is performed at loop closure and for final global relaxation. \cite{kummerle2011g} g2o library is used for graph construction and optimization. It is important to note that the focus of this paper is not the map building process, but the process of validating the map as it is used for localisation.

Maps are created in a local map coordinate system with no global reference at first. Once the map of the local area is obtained, the vehicle GPS trajectories are matched to the trajectories in local map frame. A transformation is computed by matching these trajectories. The local map is transformed into a global frame using this transformation. In this way, the relations between the local features are preserved. Global accuracy is within the range of what the ICP algorithm was able to match the feature observations against the map features(VPT metric of map-based ICP matching) using the initial position estimate from the GPS.
\subsection{Localisation}
A unscented kalman filter (UKF) fusion of GPS and IMU/encoder sensor data is used to estimate the vehicle pose until a suitable match of lidar observations to the feature map is found. At this point, the UKF performs updates using the output of the ICP global pose using the IMU/encoder to influence the filter prediction steps. The information from the GPS is not used after successful initialisation of the feature map matching due to high uncertainty and non-Gaussian nature of the GPS data in the urban environment. During GPS initialisation, the heading is calculated using the difference of two subsequent GPS readings, as this cannot be obtained from a single GPS observation. In order for heading to be relatively stable, the heading is only estimated from the GPS when the vehicle is moving above a speed threshold. The filter states are Easting, Northing, and Heading measured as an angle from East direction, as formalized in Universal Transverse Mercator(UTM) coordinate system.


\section{EXPERIMENTS AND RESULTS} \label{result}
We evaluated the metrics using the datasets described with four UKF localisation strategies:
\begin{itemize}
\item Dead-reckoning using only IMU and wheel encoders.
\item GPS filter fusing GPS and dead-reckoning odometry.
\item Pole lidar feature matching fusing ICP based pose estimation with dead-reckoning odometry. Map and lidar features are restricted to the pole class. The GPS filter is used only in the initialisation stage.
\item Pole+corner lidar feature matching fusing ICP based pose estimation with dead-reckoning odometry. Both pole and corner lidar features are utilised. The GPS filter is used only in the initialization stage.
\end{itemize}
All modes utilize the same underlying UKF filter, the observations are switched on or off depending on the selected strategy. We abbreviate these modes as Dead-reckoning, GPS, Pole and Pole+corner for simplicity. 

It is worth clarifying that the set of filter parameters for prediction and observation uncertainty used in these experiments were calculated previously and remained constant for all of the localisation datasets. 
The map is pre-existing and kept fixed for all experiments. The metrics described here are used to determine the quality of this map.
Our valid prior threshold experiments do not evaluate the map creation process, but the quality of map given a localization algorithm and robustness of localization algorithm given a pre-existing map. The purpose of these experiments is to demonstrate the usage of robustness metrics on practical localization systems and real world datasets.

In dataset 1 we drove around the area outside the Seymour Centre for six loops both clockwise, then six anticlockwise. 
Fig. \ref{fig:seymourUncerty} is a comparison of estimated uncertainty of these localisation modes in Easting, Northing and heading. The feature based localisation is far superior to the GPS pose estimates. There is an period of no observations for the feature based position estimate at around 150m, at which time the uncertainty grows to almost equal to the GPS based solution before it recovers.
In dataset 2 a 470m long road loop outside the Main Quadrangle is recorded, the online localisation trajectory mean is shown in Fig. \ref{fig:pathwithxsigma}, and standard deviation in Easting is denoted by color. It can be seen that there are areas where the feature based localisation increases in uncertainty. This is caused by a lack of global position updates in several areas. These outages can also be seen for the same trajectory in Fig. \ref{fig:allsensorEval} corresponding to the increases in uncertainty in the blue line at distances of approximately 200m and 400m.


\subsection{Probability of Absence of Updates Curves} \label{Probability of dead-reckoning result}
Fig. \ref{fig:probsLostQuad} shows the PAU graph (introduced in Section \ref{Probability of dead-reckoning}) for the 470m loop in front of the Main Quadrangle. It demonstrates the improvement of using both pole and corner features compared to using only lidar pole features. This difference can be quantitatively represented by the gap or area between the two curves. 
If we set a cut-off at 0.05 in Fig. \ref{fig:probsLostQuad} which corresponds to a 95\% likelihood of receiving at least one update, 
the longest absence of updates is 4m for both pole and corner features, and 6m for pole only features.
This means that for pole only feature matching, there is an additional 2 meters of dead-reckoning which leads to an increase in the uncertainty compared to using pole+corner features.

%
\subsection{Valid Prior Threshold (VPT) metric} \label{prior boundary result}
The valid prior threshold metric defined in Section \ref{Valid prior threshold} provides a more detailed examination of the feature matching robustness at different locations. 
We took the lidar features for each point along the trajectory and apply the valid prior threshold metric as described in Section \ref{Valid prior threshold}
The search space for the metric was set to a step size of 1m in Easting and Northing and 0.1 radian/5.7degrees in heading and the boundaries were calculated.
Fig. \ref{fig:icpBoundPerFrame} shows three examples of the matching tolerance at various different locations. In certain locations, the matching algorithm can tolerate more than +/-10m or +/-70 degrees of deviation. At some other locations there were insufficient features to make any match even with a precise pose prior.
In Fig. \ref{fig:icpBoundAll} we plot the radius of each boundary calculated along the 470m Main Quadrangle loop trajectory by computing square root of the area within the boundary at 5.7 degrees. The VPT metric radius is approximated using the median point of the area. We then plot the VPT metric radius together with 95 percent confidence bounds of the dead-reckoning, GPS filter, and lidar feature based filters in Fig. \ref{fig:allsensorEval}. Where the prior threshold is above the GPS bound, localisation is likely to be successfully initialised. Where the prior threshold is above lidar line, localisation is likely to continue successfully receiving pose updates from the ICP algorithm.
The space between the valid prior threshold and filter uncertainty is indicative of the robustness of each individual algorithm. For the Main Quadrangle map, each of the filters is considered robust with the exception of the range between 150-200m measured from the start of the trajectory. At this point, there are insufficient features meaning that dead-reckoning is required for this area and that GPS initialisation of the map matching algorithm is likely to fail.

\section{CONCLUSION AND DISCUSSION}
In this paper we introducing metrics to evaluate the robustness of localisation and feature based maps. We recognise robustness as an essential quality of localisation, and demonstrate that it can be quantified and characterised even without ground truth. Robustness has been traditionally overlooked in pursuit of more accurate and efficient algorithms, but is necessary to evaluate in order to have a more comprehensive understanding of different localisation algorithms. This paper presented a number of metrics that can be applied to any localisation algorithm operating in different environments. It can provide very useful information to detect whether a particular combination of sensors is appropriate to achieve robustness in a given domain. 

Confined by the length of this paper, only 2 short datasets are evaluated. In the future we would like to experiment on a larger volume of data, such as our weekly campus dataset, so that the results are more statistically rigorous.

\bibliographystyle{ieeetr}
\bibliography{citation}

\addtolength{\textheight}{-12cm}   

\end{document}